\pdfoutput=1
\documentclass[11pt]{article}
\usepackage{CJK}
\usepackage[]{acl}
\usepackage{tcolorbox}
\usepackage{multirow}
\usepackage{colortbl}  %彩色表格需要加载的宏包
\usepackage{xcolor}
\usepackage{multicol}
\usepackage{hyperref}
\usepackage{amsmath}
\usepackage[figuresright]{rotating}
\usepackage{color}
\usepackage{pifont}
\usepackage{enumitem}
\usepackage{booktabs}
\usepackage{amssymb}
\usepackage{adjustbox}
\usepackage{svg} % 需使用包
\usepackage{times}
\usepackage{latexsym}
\usepackage{graphicx} 
\usepackage[T1]{fontenc}

\usepackage[utf8]{inputenc}
\usepackage{microtype}
\title{PerLTQA: A Personal Long-Term Memory Dataset for Memory Classification, Retrieval, and Synthesis in Question Answering}

\author{
  Yiming Du$^{1}$, Hongru Wang$^{1}$, Zhengyi Zhao$^{1}$, Bin Liang$^{1}$, Baojun Wang$^{2}$,  Wanjun Zhong$^{2}$, \\ \textbf{Zezhong Wang}$^{1}$, \textbf{Kam-Fai Wong}$^{1*}$\\
  $^1$ The Chinese University of Hong Kong \\
  $^2$ Huawei Noah’s Ark Lab \\
  \texttt{\{ydu, kfwong\}@se.cuhk.edu.hk} \\
}

\begin{document}

\maketitle
\begin{abstract}
{
Long-term memory plays a critical role in personal interaction, considering long-term memory can better leverage world knowledge, historical information, and preferences in dialogues.
%yet its application in conversation has been largely limited to world knowledge, historical dialogues, and preferences. 
Our research introduces PerLTQA \footnote{Our code and dataset are accessible via \url{https://github.com/Elvin-Yiming-Du/PerLTQA}.}, an innovative QA dataset that combines semantic and episodic memories, including world knowledge, profiles, social relationships, events, and dialogues. This dataset is collected to investigate the use of personalized memories, focusing on social interactions and events in the QA task. PerLTQA features two types of memory and a comprehensive benchmark of 8,593 questions for 30 characters, facilitating the exploration and application of personalized memories in Large Language Models (LLMs). Based on PerLTQA, we propose a novel framework for memory integration and generation, consisting of three main components: \textbf{Memory Classification}, \textbf{Memory Retrieval}, and \textbf{Memory Synthesis}. We evaluate this framework using five LLMs and three retrievers. Experimental results demonstrate that BERT-based classification models significantly outperform LLMs such as ChatGLM3 and ChatGPT in the memory classification task. Furthermore, our study highlights the importance of effective memory integration in the QA task.
}
\end{abstract}

\begin{table*}[!t]
\centering
\renewcommand{\arraystretch}{1.05}
\resizebox{15cm}{!}{
\begin{tabular}{lcccccl}
\toprule
\multicolumn{1}{c}{\multirow{2}{*}{\textbf{Dataset}}} & \multicolumn{3}{c}{\textbf{\begin{tabular}[c]{@{}c@{}}Semantic \\ Memory\end{tabular}}} & \multicolumn{2}{c}{\textbf{\begin{tabular}[c]{@{}c@{}}Episodic \\ Memory\end{tabular}}} & \multicolumn{1}{c}{\multirow{2}{*}{\textbf{Goal}}} \\ \cline{2-6}
\multicolumn{1}{c}{} & WK & PRO & SR & DLG & EVT & \multicolumn{1}{c}{} \\ \hline
Natural-QA \citep{kwiatkowski2019natural} & \ding{51} & \ding{55} & \ding{55} & \ding{55} & \ding{55} & QA on Wikipedia \\
CoQA \citep{reddy2019coqa} & \ding{51} & \ding{55} & \ding{55} & \ding{55} & \ding{55} & Dialogue QA on world knowledge \\
HybridQA \citep{chen2020hybridqa} & \ding{51} & \ding{55} & \ding{55} & \ding{55} & \ding{55} & Multi-Hop QA on world knowledge \\
OTT-QA \citep{chen2020open} & \ding{51} & \ding{55} & \ding{55} & \ding{55} & \ding{55} & QA on tables and text \\ \hline
Multi-Woz \citep{budzianowski2018multiwoz} & \ding{55} & \ding{55} & \ding{55} & \ding{51} & \ding{55} & Task-oriented Dialogue \\
Persona-Chat \citep{zhang2018personalizing} & \ding{55} & \ding{51} & \ding{55} & \ding{51} & \ding{55} & Consistent personality dialogue \\
DailyDialog \citep{li2017dailydialog}& \ding{55} & \ding{55} & \ding{55} & \ding{51} & \ding{55} & Multi-turn dialogues on daily life \\
Personal-Dialogue \citep{zheng2019personalized} & \ding{55} & \ding{51} & \ding{55} & \ding{51} & \ding{55} & Multi-turn personalized dialogues \\
MSC \citep{xu2021beyond} & \ding{55} & \ding{51} & \ding{55} & \ding{51} & \ding{55} & Long-Term open-domain conversation \\
DialogueSum \citep{chen2021dialogsum} & \ding{55} & \ding{55} & \ding{55} & \ding{51} & \ding{55} & Dialogue summarization \\
Dulemon \citep{xu2022long} & \ding{55} & \ding{51} & \ding{55} & \ding{51} & \ding{55} & Personal long-term Chinese conversation \\
HybridDialogue \citep{nakamura2022hybridialogue}& \ding{51} & \ding{55} & \ding{55} & \ding{55} & \ding{55} & Dialogue QA on tables and text \\
Topical-Chat  \citep{gopalakrishnan2023topical} & \ding{51} & \ding{55} & \ding{55} & \ding{55} & \ding{55} & Knowledge-grounded open-domain conversations \\
ChatDB \citep{hu2023chatdb} & \ding{51} & \ding{55} & \ding{55} & \ding{55} & \ding{55} & Question answering with structured memory \\
MemoryBank \citep{zhong2023memorybank} & \ding{55} & \ding{51} & \ding{55} & \ding{51} & \ding{55} & Personal long-term memory dialogue \\ \hline

PerLTQA  & \ding{51} & \ding{51} & \ding{51} & \ding{51} & \ding{51} & \multicolumn{1}{l}{\begin{tabular}[l]{@{}l@{}}Question answering on personal long-term memory \\ including semantic and episodic memory\end{tabular}} \\ 
\bottomrule
\end{tabular}
}
\caption{Typology of memories in QA/Dialogue datasets: Analysis of World Knowledge (WK), Profiles (PRO), Social Relationships (SR), Dialogues (DLG), and Events (EVT).}
\label{long_term_mem_dataset}
\vspace{-0.5cm}
\end{table*}

\section{Introduction}
\begin{figure*}[t]
\centering
\includegraphics[width=0.95\textwidth]{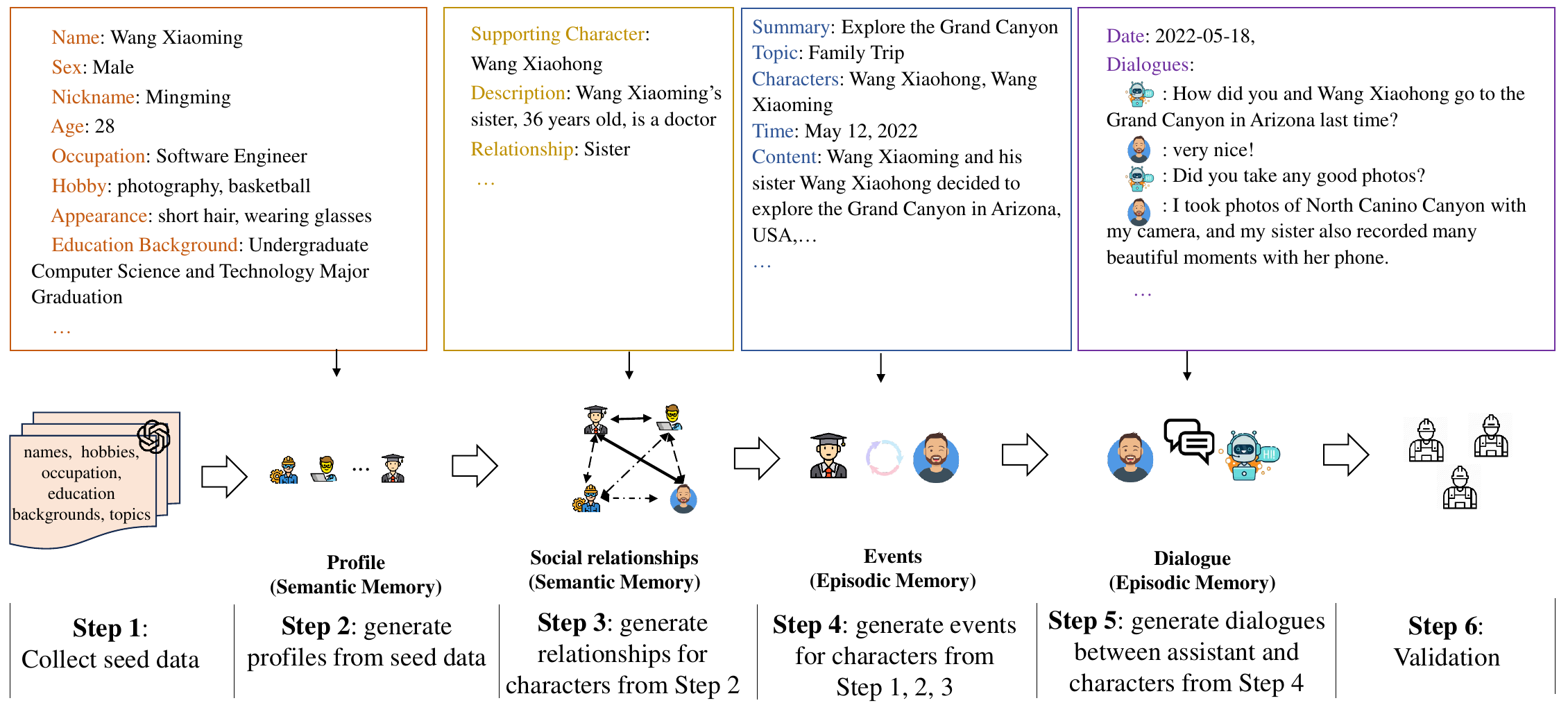}
\captionsetup{justification=justified}
\caption{The process of PerLT Memory generation. A six-step process: Step 1. Seed data collection. Step 2. PRO generation. Step 3. SR generation. Step 4. EVT generation. Step 5. DLG generation and Step 6. Validation.}
\label{fig1:variamem_dataset_generation}
\vspace{-0.5cm}
\end{figure*}

{
Long-term memory is an essential component of the human memory system, characterized by the ability to store extensive information and retrieve it when necessary. Spanning durations from mere minutes to an entire lifetime, this type of memory is fundamental to cognitive functions, as noted by \citep{tulving2000oxford}. Within the conversational domain, as explored by \citep{xu_beyond_2021, zhong2023memorybank}, integrating personal long-term memories can yield more personalized responses. Consequently, simulating the mechanisms of long-term memory is important for improving current dialogue systems.

Cognitive science classifies long-term memory into episodic and semantic types \citep{atkinson1968human}. \textbf{Semantic memory} \citep{eysenck2020cognitive} is a mental representation that involves personal facts and world knowledge, such as profiles and relationships. It does not depend on the specific experience of the individual. While \textbf{episodic memory} \citep{eysenck2020cognitive} is about personal histories, specifically in events and dialogues. Previous research indirectly employs data aligning with episodic and semantic memory categories of cognitive psychology, despite not explicitly adopting its memory frameworks \citep{eysenck2020cognitive}. Traditional QA systems \citep{kwiatkowski2019natural, reddy2019coqa, chen2020hybridqa} and dialogue systems \citep{nakamura2022hybridialogue, gopalakrishnan2023topical, hu2023chatdb} initialize semantic memory as world knowledge from an external database. Existing dialogue systems consider dialogue history \citep{budzianowski2018multiwoz, zhang2018personalizing, li2017dailydialog, zheng2019personalized, xu2021beyond} or dialogue summaries \citep{xu2021beyond, chen2021dialogsum} as episodic memory. Research on personalization within dialogue systems \citep{zhang2018personalizing, zheng2019personalized, xu2022long} has increasingly concentrated on the preferences in dialogues, uncovering their substantial influence on personal memory. We consolidate these findings into Table \ref{long_term_mem_dataset} and identify a notable research gap in integrating personal semantic memory concerning profiles, social relationships and episodic memory concerning events into the QA system.

To explore personal long-term memory in question answering \citep{nakamura2022hybridialogue}, we propose the PerLTQA dataset, as depicted in Figure \ref{fig1:variamem_dataset_generation}. The PerLTQA dataset is designed to capture the essence of semantic and episodic memories. It is constructed using an in-context approach \citep{brown2020language} that merges personal memory integration with contextual generation. This process produces an extensive representation of character-specific long-term memory, encompassing profiles (semantic memory), social relationships (semantic memory), events (episodic memory), and dialogues (episodic memory) as shown in Figure \ref{fig1:variamem_dataset_generation}. PerLTQA features various QA items, including questions, answers, reference memories, and memory anchors (the key memory fragments in the answer are targeted to the question). The memory anchors and reference memory are used for aligning answers with their corresponding memories during evaluations. Our methodology involves three steps: \textbf{memory classification}, \textbf{memory retrieval}, and \textbf{memory synthesis}. Initially, we classify the given question to determine its memory type, ensuring an accurate categorization of memory. Next, we proceed to memory retrieval, where we utilize the probabilities derived from the classification stage to re-rank and prioritize the retrieved memories related to the question. Finally, in the memory synthesis stage, we generate the answers in LLMs by integrating the re-ranked memory information.

In summary, our research makes three main contributions: 

\begin{itemize}[leftmargin=*,topsep=2pt,itemsep=2pt,parsep=0pt]
    \item Our research contributes to introducing the PerLTQA dataset, encompassing a memory database with 141 profiles, 1,339 semantic social relationships, 4,501 events and 3,409 dialogues, and 8,593 memory-related evaluation questions.

    \item  We propose three subtasks memory classification, memory retrieval, and memory synthesis to evaluate the memory utilization capabilities of LLMs. We carry out experiments using five LLMs and three retrieval models.

    \item Our experiment results indicate that the BERT \citep{devlin2018bert} model excels in memory classification tasks, surpassing LLMs like ChatGLM3 \citep{zhang2023glm} and ChatGPT \footnote{https://chat.openai.com.}, and LLMs show varied proficiency in generating memory-based responses when provided with accurately retrieved memories.
\end{itemize}
}

\section{Related Work}
In cognitive psychology, semantic memory relates to world knowledge and social relationships, whereas episodic memory involves events. This differentiation is mirrored in the datasets like \citep{kwiatkowski2019natural, chen2021dialogsum, zhong2023memorybank}. In the realm of question answering \citep{kwiatkowski2019natural, reddy2019coqa, chen2020hybridqa, chen2020open}, Natural-QA \citep{kwiatkowski2019natural} and CoQA \citep{reddy2019coqa} both target Wikipedia-based knowledge, exemplifying the use of world knowledge as semantic memory. Within dialogue tasks \citep{wang2023survey}, MSC \citep{xu2021beyond} and Dulemon \citep{xu2022long} consider dialogues as episodic memory. MemoryBank \citep{zhong2023memorybank} introduces a bilingual dataset using GPT-4 to summarize dialogues and personal data, effectively simulating episodic memory in multi-turn dialogues. However, existing datasets \citep{hu2023chatdb, zhang2023memory} lack comprehensive coverage of both memory types with detailed annotations on social relationships and events, highlighting a research gap for LLMs in personal long-term memory synthesis.

Memory retrieval is a crucial component for QA and dialogue system to successfully generate memory-based responses. The existing retrieval methods fall into three main categories: BM25 \citep{robertson1995okapi}, employing a statistical approach for document ranking based on query terms; DPR \citep{karpukhin2020dense}, demonstrating supervised retrieval capabilities; and Contriever \citep{izacard2021unsupervised}, showcasing unsupervised retrieval techniques. With the rise of LLMs \citep{wang2023survey}, an increasing number of works \citep{zhang2023memory, zhong2023memorybank} are utilizing the RAG (Retrieval-Augmented Generation) \citep{lewis2020retrieval} for retrieval-enhanced tasks. Within this framework, fine-tuned embeddings are employed for text similarity searches, such as REPLUG \citep{shi2023replug}, OpenAI Embeddings \footnote{https://platform.openai.com/docs/api-reference/embeddings}.  This concept has been implemented in frameworks such as LlamaIndex \footnote{https://docs.llamaindex.ai/en/latest/index.html}, LangChain \footnote{https://www.langchain.com/}. This approach leverages the refined embeddings to efficiently identify and retrieve content that is most relevant to the given query. 

Aiming to integrate retrieved memories into responses, LLMs provide a prompt-based generation method facilitating the generation of memory-informed responses \citep{zhang2023glm, yang2023baichuan, bai2023qwen, zhang2023internlm, touvron2023llama2}. In the dialogue system \citep{zhao_unimc_2023, lee_prompted_2023, zhong2023memorybank}, they incorporate memory into the prompt and generate the memory-related response. In this way, we can improve the relevance and specificity of the response, leading to context-aware responses of personal memory. 

\section{Dataset Collection}
We detail the creation of the PerLTQA dataset, which involves collecting PerLT memories and generating and annotating PerLT QA pairs. Using in-context technique, we build a memory database and semi-automatic annotate memory-based Q\&A pairs.

\subsection{PerLT Memory Generation}
As shown in Figure \ref{fig1:variamem_dataset_generation}, the generation of PerLT memories is decomposed into six steps: 

\noindent \textbf{Step 1. Diverse Seed Data Collection.} We select ChatGPT and Wikipedia as initial sources for our seed dataset due to their comprehensive coverage of a wide range of occupations, educational backgrounds, hobbies, and event topics, essential for foundational knowledge. It comprises professional backgrounds that span across 10 categories and 299 specialties, hobbies that are categorized into 7 groups with 140 items, and a comprehensive range of topics structured into 49 categories with 2442 subtopics. Complementing this approach, \texttt{gpt-3.5-turbo} is employed to generate 141 virtual names. We implement a manual review process, allowing us to avoid the unrealistic use for data generation.

\noindent \textbf{Step 2. Profile (Semantic Memory) Generation.} To study personalized memories, generating character profiles is essential. We leverage seed data, particularly occupations, educational backgrounds, hobbies inputs, within prompt templates that include descriptions of other attributes (gender, nickname, age, nationality, appearance, achievements, education, profession, employer, awards, and role models). By utilizing ChatGPT (\texttt{gpt-3.5-turbo}), we generate random character profiles. The detailed prompts for this process is available in Appendix.\ref{memory_generation_prompt}.

\noindent \textbf{Step 3. Social Relationship (Semantic Memory) Generation.} For the development of diverse social connections, we utilize structured prompts shown in Appendix.\ref{memory_generation_prompt} to craft 50 distinct categories of relationships. These categories span a wide array, including but not limited to family, friends, romantic partners, acquaintances, colleagues, mentor/student dynamics, and neighbors, aiming to comprehensively cover social interactions.

\noindent \textbf{Step 4. Event (Episodic Memory) Generation.} Each character includes a series of narrative events, deeply embedded in their episodic memory and linked to interactions with others. The event generation starts by generating descriptions of background events chosen at random from the seed topics highlighted in Step 1. Following this step, we use prompts to help create detailed accounts of events that are deeply tied to these initial occurrences and the web of social connections. To ensure coherence between the dynamics of character interactions and the backdrop of events, few-shot learning techniques, as outlined by \citep{brown2020language}, are employed. This strategy aids ChatGPT (\texttt{gpt-3.5-turbo}) in achieving narrative consistency, weaving together individual events and relationships into a cohesive story for each character.

\noindent \textbf{Step 5. Dialogues (Episodic Memory) Generation.} 
Building on the events generated in Step 4, we craft historical dialogues between the AI assistant and the character. This process, anchored in historical events, ensures that conversations maintain relevance to past occurrences. We utilize prompt templates that merge character profiles and event details to help dialogue generation, as detailed in Appendix.\ref{memory_generation_prompt}. Furthermore, embedding the dialogues maintains a profound connection to the shared histories and relationships.

\noindent \textbf{Step 6. Validation.} We start with small batches for quality checks and scale up after ensuring error-free outputs. We conduct random sampling of the generated memory data, identifying types of issues as detailed in Appendix \ref{memory_error_types}, and then manually refine the memories. This refinement includes removing anomalies in profiles, discriminatory content, inconsistencies in character memories, and brief event narratives, enhancing the accuracy and consistency of the memory.

\subsection{PerLT Question Answering}
{
To thoroughly assess each memory type for a character, we gather four QA-related metrics (\textit{question}, \textit{answer}, \textit{reference memory}, and \textit{memory anchor}) for evaluating the memory-based QA. The process of collecting PerLT QA items unfolds in three phases:

\noindent \textbf{Question and answer generating.} Utilizing ChatGPT, we generate questions and answers prompted by the memory sentences stored in PerLT Memory database. The answers are designed to align with the reference memories provided, adhering to the prompts we created, as shown in the Appendix.\ref{memory_qa_prompt}. 

\noindent \textbf{Memory Anchor Annotation.} The memory anchor, a key text segment in the answer that aligns with the referenced memory and question, is essential for memory evaluation in response generation. We employ exact match techniques and human verification to annotate the start and end positions of memory anchors, guided by the reference memory. Given the intensive labor involved in manual adjustments, we have annotated memory anchors for a limited set of 30 characters.

\noindent \textbf{Validation on QA pairs and Memory Anchor.} To ensure the integrity of PerLT QA pairs, we initiate our quality control with an unbiased random sampling. This is followed by a detailed categorization of errors in QA, references, and memory anchors, supplemented by thorough pronominal reference checks for accuracy. All error types are meticulously cataloged in the Appendix.\ref{sec: appendix}. We use LLMs as a scoring mechanism, evaluating on a scale from 0 to 10, directly accepting QA pairs with a score of 10, reviewing those scored between 7 and 9, and eliminating those below 6. Automated validation is conducted to ensure the accuracy of reference memories and to remove irrelevant stop words. This is followed by meticulous manual corrections and alignment checks between memory anchors and references, ensuring the highest quality of QA items.}

\subsection{Dataset Statistics}

\begin{table}[!t]
\centering
\renewcommand{\arraystretch}{1.05}
\resizebox{8cm}{!}{
\begin{tabular}{lll}
\toprule
\multicolumn{3}{l}{Dataset Statistics}                                                                                                     \\ \hline
\multirow{2}{*}{Profiles}  & \# Character profiles & 141    \\
& \# Jobs    & 98     \\ \hline
\multirow{3}{*}{\begin{tabular}[c]{@{}c@{}}Semantic \\ Memory\end{tabular}}   & \# Relationship Descriptions   & 1,339  \\
& \# Relationship Categories  & 50     \\
& \begin{tabular}[c]{@{}l@{}}\# Average Social Relationships\\  per Character\end{tabular} & 9.5    \\ \hline
\multirow{7}{*}{\begin{tabular}[c]{@{}c@{}}Episodic \\ Memory\end{tabular} }    & \# Topics  & 49    \\
& \# Events  & 4,501  \\
& \# Average Words per Events   & 313    \\
& \# Event-related Historical Dialogs  & 3,409  \\
& \# Utterances  & 25,256 \\
& \# Average Words per Utterance & 43.7   \\ \hline
\multirow{4}{*}{\begin{tabular}[c]{@{}c@{}}Memory \\ QA\end{tabular} } & \# Question Answer Pairs  & 8,593  \\
& \# Average Words per Question   & 16.7   \\
& \# Average Words per Answer  & 27.4   \\
& \# Memory Anchors  & 23,697 \\
& \# Average Anchors & 2.8    \\ 
\bottomrule
\end{tabular}
}
\caption{PerLTQA dataset statistics. }
\label{PerLTQA_dataset}
\vspace{-0.5cm}
\end{table}

{The PerLTQA dataset, presented in Table \ref{PerLTQA_dataset}, includes 141 character profiles with detailed occupations and relationships. With 50 relationship categories, an average of 9.5 social relationships per character, the dataset provides a vavid social relationship for semantic memory. Furthermore, PerLT Memory features 4,501 events, averaging 313 words each, which fuel 3,409 event-related historical dialogues, totaling 25,256 utterances. In the QA section, 8,593 question-answer pairs and 23,697 memory anchors average 16.7 and 27.4 words, respectively. This rich compilation of data supports the development of dialogue QA system with a profound understanding of human-like memory recall and synthesis within a concise framework.
}

\begin{figure*}[t]
\centering
\includegraphics[width=0.75\textwidth]{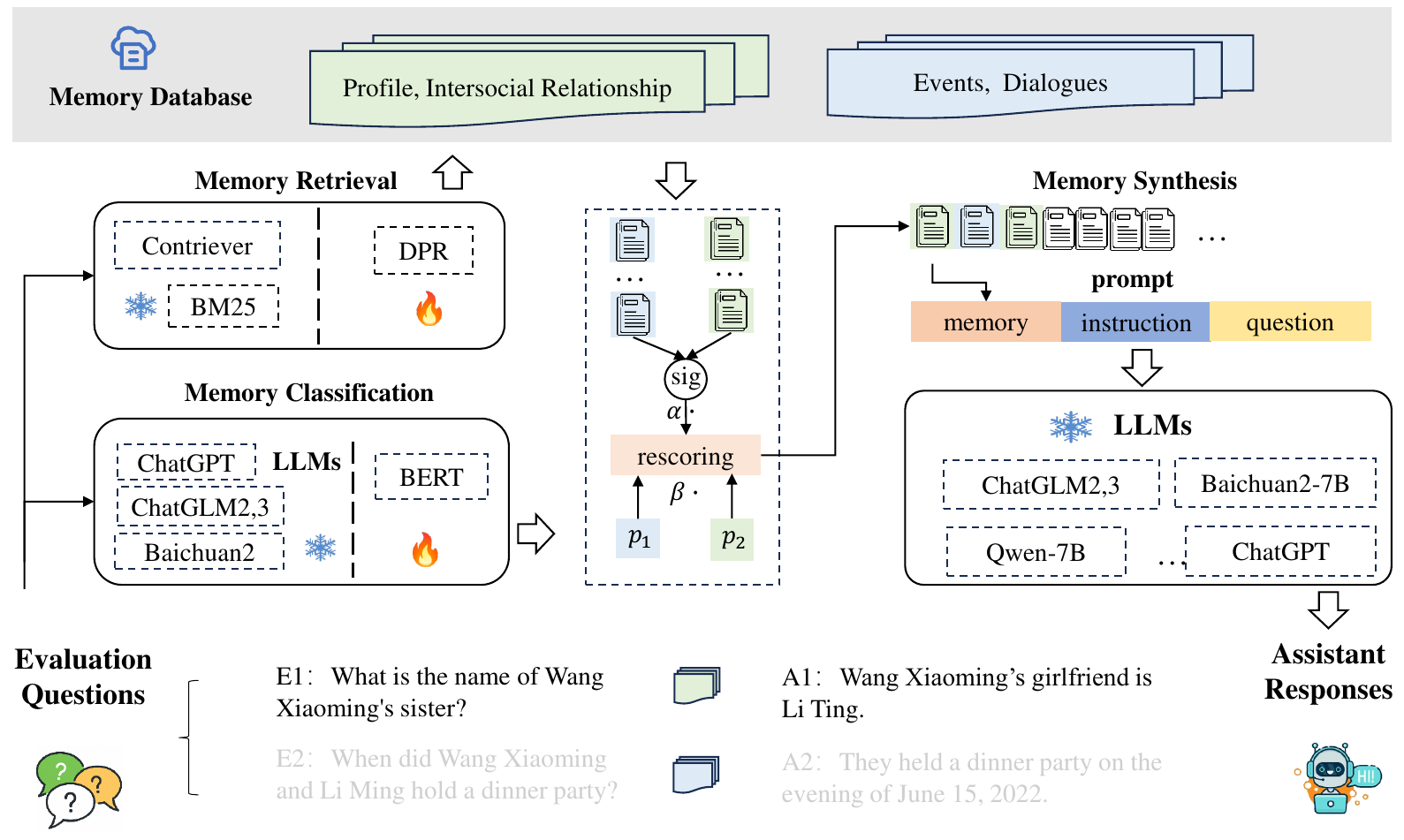}
\captionsetup{justification=justified}
\caption{The framework of memory classification, memory retrieval and memory synthesis in QA.}
\label{fig2:framework}
\vspace{-0.6cm}
\end{figure*}

\subsection{Task Definition}
\label{task_definition}
The PerLT memory database is formulated as $M = \left\{ (S_i(l_1), E_i(l_2)) \mid i = 1, 2, \ldots, p \right\}$, where each tuple consists of semantic memories including profiles and social relationship and episodic memories including events and dialogs. Each $S_i(l_1)$ and $ E_i(l_2)$ are defined to have $l_1$, $l_2$ elements, respectively, which are specific to the i-th character memory representation.

The PerLT QA dataset comprises a set of items \( T = \{ t_j \}_{j=1}^N \), where each item \( t_j \) is a tuple consisting of four elements: \( t_j = (q_j, r_j, m_j, a_j) \). Here, \( q_j \) denotes the question, \( r_j \) the reference memory, \( m_j \) the memory anchor, and \( a_j \) the answer. The dataset spans various data types including semantic memory, and episodic memory, which are implicitly reflected in the construction of each \( t_j \). The variable \( N \) represents the total number of QA items in the dataset.

% \section{Tasks and Baseline Models}
As shown in Figure \ref{fig2:framework}, to explore the integration of memory information in QA, we propose three subtasks: \textit{memory classification}, \textit{memory retrieval} and \textit{memory synthesis} for response generation. In particular, memory synthesis is our ultimate goal.

\textbf{Memory Classification.} We introduce a classification model designed to assist queries in finding semantic memory or episodic memory. This model can operate through an instruction-based LLM, few-shot-based LLM, or BERT-based classifier. The classification model conforms to a unified formula as Eq.(\ref{eq1})
\begin{equation}
\pi = MC (q)
\label{eq1}
\end{equation}
where $\pi$ represents the classification result, $MC$ denotes the classification model and $q$ is the input query. The outputs from our classification model enhance memory retrieval by facilitating post-ranking of the retrieved memories, thereby reducing the excessive dependence on memory classification within the framework. Further details are elaborated in Appendix.\ref{opt_rerank}.

\textbf{Memory Retrieval.} We aim to perform memory retrieval by extracting relevant character memories for a given evaluation question from the PerLT memory database $M$, formalized as Eq.(\ref{eq2}).
\begin{equation}
m, s  =  R(q, M, k)
\label{eq2}
\end{equation}
where $m$ is the retrieved memory with size $k$, $s$ is the corresponding scores, $R$ is the retrieval model.

Our method distinguishes itself by initially retrieving k memories from each category within the memory database, amassing $2k$ potential memory candidates. These candidates undergo a re-ranking process influenced by their classification scores, culminating in a composite score for each memory $m_i$, which is computed as follows:
\begin{equation}
s'_{i} = \alpha \cdot P(\pi|m_i) + \beta \cdot \mathrm{sigmoid}(s_{i})
\end{equation}
where $P(\pi|m_i)$ is the probability given by the classification model that the memory item \(m_i\) belongs to $\pi$. The top $k$ memories are then selected based on these final scores. $\alpha$ and $\beta$ represent the weight of each term, and we set both to 0.5 to balance their contributions.

\textbf{Memory Synthesis.} Memory synthesis leverages $LLM$ for response generation. This task uses a prompt template $z$ (as illustrated in Appendix.\ref{t_prompt}), an evaluation question $q$, and retrieved memories $m$ as Eq.(\ref{eq3}).

\begin{equation}
r' = LLM ( z, q, m )
\label{eq3}
\end{equation}

\subsection{Evaluation Metrics}
{
For the memory classification task, we use precision (P), recall (R), F1, and Accuracy to serve as metrics. For the memory retrieval task, we utilize Recall@K \citep{manning2008introduction} as our metric. To evaluate memory synthesis for the response generation task, we measure the correctness and coherence of responses with \texttt{gpt-3.5-turbo}-based evaluation method \citep{zhong2023memorybank} and use MAP (mean average precision) of memory anchors as shown in Eq.(\ref{mmap_equ}) to evaluate memory synthesis ability \citep{nakamura2022hybridialogue}.}
\begin{equation}
\label{mmap_equ}
\text{MAP} = \frac{1}{N} \sum_{i=1}^{N} \frac{\mathrm{EM}(q_i, mar_i)}{\mathrm{NUM}(mar_i)}
\end{equation}

where N represents the total number of questions in the evaluation dataset. $mar$ denotes memory anchors, $\mathrm{EM}$ represents the tally of exact matches between queries and memory anchors, and $\mathrm{NUM}(mar_i)$ is the count of memory anchors per question.

\section{Experiments}

\begin{table*}[]
\centering
\renewcommand{\arraystretch}{1.05}
\captionsetup{justification=justified}
\resizebox{13cm}{!}{
\begin{tabular}{l|lll|lll|lll}
\toprule
\multirow{2}{*}{} & \multicolumn{3}{c}{\textbf{W-MC+R}} & \multicolumn{3}{|c}{\textbf{W/o-MC+W-R}} & \multicolumn{3}{|c}{\textbf{W/o-MC+R}} \\ \cline{2-10} 
 & \multicolumn{1}{c}{\textbf{MAP}} & \multicolumn{1}{c}{\textbf{Corr.}} & \multicolumn{1}{c}{\textbf{Coh.}} & \multicolumn{1}{|c}{\textbf{MAP}} & \multicolumn{1}{c}{\textbf{Corr.}} & \multicolumn{1}{c}{\textbf{Coh.}} & \multicolumn{1}{|c}{\textbf{MAP}} & \multicolumn{1}{c}{\textbf{Corr.}} & \multicolumn{1}{c}{\textbf{Coh.}} \\ \hline
ChatGLM2 & 0.688 & 0.483 & 0.963 & 0.688 & 0.481 & 0.962 & 0.128 & 0.054  & 0.960 \\
ChatGLM3 & 0.704 & 0.517 & 0.971 & 0.695 & 0.517 & 0.969 & 0.130 & 0.060 & 0.962 \\
Qwen-7B & 0.729 & 0.535 & 0.960 & 0.720 & 0.532 & 0.959 & 0.131 & 0.057 & 0.957 \\
Baichuan2-7B & 0.736 & 0.535 & 0.966 & 0.728 & 0.522 & 0.968 &  0.132 & 0.051 & 0.953 \\
gpt-3.5-turbo & 0.756 & 0.573 & 0.969 & 0.745 & 0.562 & 0.969 & 0.156 & 0.088 & 0.961 \\ 
\bottomrule
\end{tabular}
}
\caption{Comparison of MAP, Correctness (Corr.), Coherency (Coh.) across three settings: With memory classification and retrieval (W-MC+R), without memory classification but with retrieval (W/o-MC+W-R), and without memory classification and without retrieval (W/o-MC+R).}
\label{memory_incorporation_res}
\end{table*}

\subsection{Implementation details}
{
In our work, we divide the data from the PerLT QA dataset into training (5155), validation (1719), and test sets (1719) for model training and evaluation. In the memory classification task, we fine-tune BERT-base model and compare the sentence classification results on the test dataset with ChatGLM2, ChatGLM3 \citep{zhang2023glm}, Baichuan2-7B-Chat \citep{yang2023baichuan}, Qwen-7B-Chat \citep{bai2023qwen}, and ChatGPT under instructional and few-shot settings. For the memory retrieval task, we employ three retrieval models - DPR \citep{karpukhin2020dense}, BM25 \citep{robertson1995okapi}, and Contriever \citep{izacard2021unsupervised} - to collect character memories. In the memory synthesis task, we use the above five LLMs to generate responses of no more than 50 words, given re-ranked retrieved memories, employing in-context learning methods. 

The memory synthesis task is evaluated across three scenarios: with memory classification and retrieval (W-MC+R), without memory classification but with retrieval (W/o-MC+W+R), and without both classification and retrieval (W/o-MC+R). Experiment details are shown in the appendix.\ref{exp_settings}
}

\begin{table}[!t]
\centering
\renewcommand{\arraystretch}{1.05}
\begin{adjustbox}{max width=0.45\textwidth}
    
\begin{tabular}{lrrrr}
\toprule
\multicolumn{1}{c}{\textbf{Metrics}} & \multicolumn{1}{c}{\textbf{P}} & \multicolumn{1}{c}{\textbf{R}} & \multicolumn{1}{c}{\textbf{F1}} & \multicolumn{1}{c}{\textbf{Acc}} \\ \hline
ChatGLM2-6B & 0.749 & 0.712 & 0.729 & 0.712 \\
ChatGLM3-6B & 0.864 & 0.485 & 0.538 & 0.485 \\
Qwen-7B & 0.730 & 0.631 & 0.673 & 0.631 \\
Baichuan2-7B & 0.848 & 0.602 & 0.657 & 0.602 \\
gpt-3.5-turbo & 0.868 & 0.668 & 0.715 & 0.668 \\ \hline
F+ChatGLM2-6B & 0.770 & 0.806 & 0.785 & 0.806 \\
F+ChatGLM3-6B & 0.778 & 0.445 & 0.508 & 0.445 \\
F+Qwen-7B & 0.804 & 0.402 & 0.452 & 0.402 \\
F+Baichuan2-7B & 0.860 & 0.324 & 0.337 & 0.324 \\
F+gpt-3.5-turbo & 0.864 & 0.511 & 0.566 & 0.511 \\ \hline
I+BERT-base & 0.720 & 0.849 & 0.779 & 0.849 \\
BERT-base & \textbf{0.960} & \textbf{0.956} & \textbf{0.957} & \textbf{0.956} \\ \bottomrule
\end{tabular}
\end{adjustbox}
\captionsetup{justification=justified}
\caption{Comparative performance of five LLMs and BERT in memory classification tasks under few-shot settings (F) and instruction-based training (I).}
\label{memory_classification_res}
\end{table}

\begin{table}[]
\centering
\renewcommand{\arraystretch}{1.05}
\begin{adjustbox}{max width=0.45\textwidth}
\begin{tabular}{l|cccc|c}
\toprule
 \multicolumn{1}{c|}{\textbf{RM}}& \multicolumn{1}{c}{\textbf{R@1}} & \multicolumn{1}{c}{\textbf{R@2}} & \multicolumn{1}{c}{\textbf{R@3}} & \multicolumn{1}{c}{\textbf{R@5}} & \multicolumn{1}{|c}{\textbf{T(s)}}\\ \hline
Contriever & 0.486 & 0.674 & 0.737 & 0.792 & 0.070 \\
DPR & 0.602 & 0.803 & 0.862 & \textbf{0.919} & 2.960 \\
BM25 & \textbf{0.705} & \textbf{0.847} & \textbf{0.871} & 0.895 & \textbf{0.030} \\\bottomrule
\end{tabular}
\end{adjustbox}
\captionsetup{justification=justified}
\caption{Performance of Recall@K (R@K) and average retrieval time (T) in memory retrieval using Contriever, BM25, and DPR models.}
\label{retrieval_res}
\vspace{-0.5cm}
\end{table}

\subsection{Memory Classification}
\label{memory_planner_sec}
{\textbf{BERT-based model provides better performance than LLMs for memory classification.} As shown in Table \ref{memory_classification_res}, BERT demonstrates superior performance compared to other LLMs under instruction and few-shot settings. Specifically, in few-shot scenarios where an evaluation question is paired with corresponding examples for each type of memory, the performance of \texttt{gpt-3.5-turbo} declines in comparison to methods that rely solely on instruction-based classification. In summary, the BERT-base model achieves the highest weighted precision (95.96\%), weighted recall (95.64\%), weighted F1 score (95.74\%), and accuracy (95.64\%). Moreover, the high performance in memory classification reinforces confidence in the rescoring mechanism, as illustrated in Figure \ref{fig2:framework}.
}

\subsection{Memory Retrieval}
\textbf{Different retrieval models show variable Recall@K and time performance.} In the memory retrieval task, Table \ref{retrieval_res} reveals that the unsupervised retrieval model Contriever significantly lags behind the statistic-based BM25 and the supervised DPR model. Moreover, as the top k values increase, DPR notably improves Recall@K performance, surpassing BM25 after k equals 3. However, the retrieval time cost of DPR is substantially higher than BM25 retrieval. This suggests that we need to balance the retrieval performance and time cost when deployment in dialogue QA tasks.
\vspace{-0.2cm}
\subsection{Memory Synthesis}
\textbf{Memory classification and retrieval significantly improve LLMs to integrate memory into responses.} The results shown in Table \ref{memory_incorporation_res} indicate that LLMs augmented with memory classification and retrieval model show a marked improvement in generating memory-informed responses over those relying solely on LLMs itself, with notable increases in precision (MAP peaking at 0.756) and correctness (correctness reaching up to 0.573). The absence of memory classification (W/o-MC+W-R) has a minimal impact on improving MAP and Correctness, maintaining robust scores (0.688-0.745 for MAP), which emphasizes the importance of retrieval mechanisms in sustaining performance. Moreover, coherency levels are impressively stable across all configurations, never falling below 0.953, reflecting the strength of LLMs in producing coherent text. These results highlight the crucial importance of the retrieval model and maintaining coherence throughout the process. Additionally, models with fewer than 10 billion parameters have shown memory synthesis capabilities similar to those of ChatGPT, indicating that smaller models can also be optimized to produce outputs of comparable quality.

\section{Analysis and Case Study}

\subsection{Ablation Study}

\begin{table}[!t]
\centering
\renewcommand{\arraystretch}{1.05}
\begin{adjustbox}{max width=0.48\textwidth}
\begin{tabular}{llccccc}
\toprule
\multirow{2}{*}{\textbf{Models}} & \multicolumn{2}{c}{\textbf{NR}} & \multicolumn{2}{c}{\textbf{IR}} & \multicolumn{2}{c}{\textbf{CR}} \\
 & \multicolumn{1}{c}{\textbf{MAP}} & \textbf{Corr.} & \textbf{MAP} & \textbf{Corr} & \textbf{MAP} & \textbf{Corr.} \\ \toprule
Baichuan2-7B & 0.132 & 0.051 & 0.396 & 0.225 & 0.782 & 0.581\\
Qwen-7B & 0.131 & 0.057 & 0.390 & 0.221 & 0.786 & 0.574\\
ChatGLM2 & 0.128 & 0.054 & 0.396 & 0.248 & 0.738 & 0.523\\
ChatGLM3 & 0.130 & 0.060 & 0.365 & 0.216 & 0.754 & 0.561\\
ChatGPT & 0.156 & 0.088 & 0.375 & 0.252 & 0.842 & 0.609\\
\bottomrule
\end{tabular}
\end{adjustbox}
\captionsetup{justification=justified}
\vspace{-0.2cm}
\caption{Performance of LLMs on MAP and Correctness (Corr.) under No Retrieval (NR), Incorrect Retrieval (IR) and correct retrieval (CR) settings.}
\label{inc_c_res}
% \vspace{-0.5cm}
\end{table}

\begin{table}[]
\centering
\renewcommand{\arraystretch}{1.05}
\begin{adjustbox}{max width=0.48\textwidth}
\begin{tabular}{lllllll}
\toprule
\multirow{2}{*}{} & \multicolumn{3}{l}{\textbf{Semantic Memory}} & \multicolumn{3}{l}{\textbf{Episodic Memory}} \\ \cline{2-7} 
 & \textbf{MAP} & \textbf{Corr.} & \textbf{Coh.} & \textbf{MAP} & \textbf{Corr.} & \textbf{Coh.} \\ \hline
gpt-3.5-turbo & 0.242 & 0.150 & 0.834 & 0.721 & 0.543 & 0.966 \\ \bottomrule
\end{tabular}
\end{adjustbox}
\captionsetup{justification=justified}
\caption{The results of gpt-3.5-turbo performance solely on episodic or semantic memory on the metric of MAP, Correctness, and Coherency}
\label{type_enhance}
\vspace{-0.5cm}
\end{table}

\textbf{Correct memory retrieval significantly enhances the accuracy of responses across various LLMs.} The experimental results, as shown in Table \ref{inc_c_res}, demonstrate the consistent ability of different LLMs to generate responses that are both relevant and accurate.  This consistency underscores that LLMs experience a substantial improvement when they have access to accurate external memory. The findings further indicate that LLMs possess a degree of tolerance towards misinformation and are capable of leveraging accurate memory information to some extent. Despite incorrect memory retrieval, all models manage to sustain a reasonable degree of precision, with MAP scores from 0.365 to 0.396, underlining their robustness in less-than-ideal information conditions.

\textbf{Semantic and episodic memory contribute to improving memory synthesis.}
As shown in Table \ref{type_enhance}, the results illustrate the comparative performance of gpt-3.5-turbo when integrating only with semantic or episodic memory. The provision of only one type of memory—either episodic or semantic, leads to varying levels of performance degradation. This variation can be attributed to the different proportions of episodic and semantic memory data within the test set, which in turn affects the response accuracy of the Large Language Models (LLMs). However, integrating any form of personal long-term memory, whether episodic or semantic, into LLMs proves beneficial for generating responses that are informed by past personal history. This suggests that the presence of personal long-term memory, regardless of its type, help LLMs to produce relevant and accurate responses.

\subsection{Case Study}
\begin{figure}[t]
\centering
\includegraphics[width=0.48\textwidth]{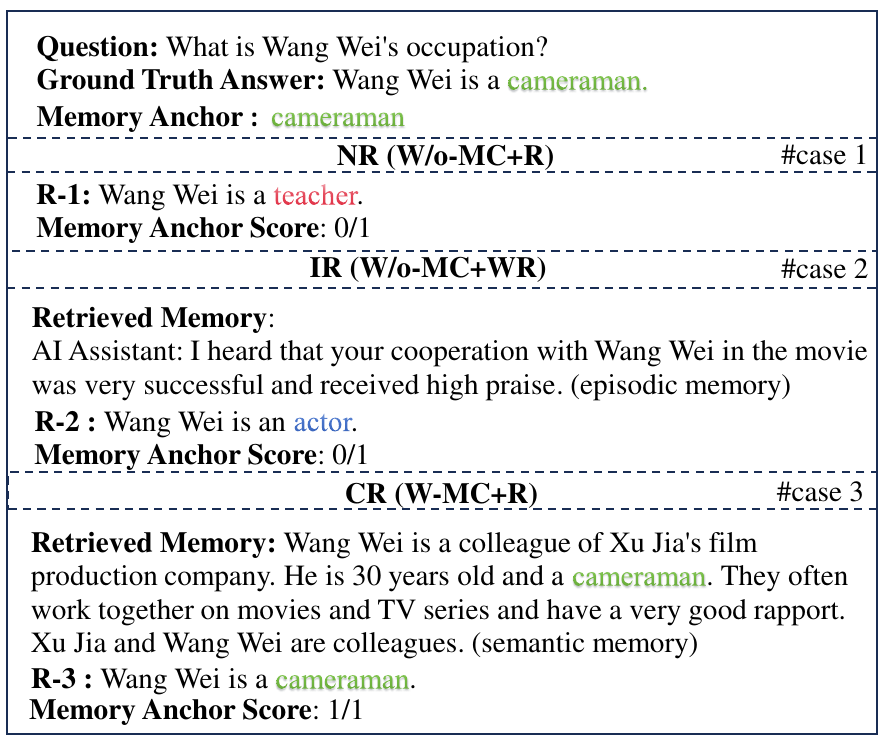}
\captionsetup{justification=justified}
\vspace{-0.3cm}
\caption{Comparative analysis of response performance without retrieval (NR), incorrect retrieval (IR), and Correct Retrieval (CR).}
\label{fig4:case study}
\vspace{-0.5cm}
\end{figure}

We present specific cases in Figure \ref{fig4:case study} to evaluate the question `What is Wang Wei's occupation?' with the verifiable answer 'cameraman'. Without memory retrieval, \textit{gpt-3.5-turbo} generates a speculative response 'Wang Wei is a teacher', a common hallucination in most LLMs, or provides context-less responses. Introducing memory retrieval, we observe two cases. In case 2, the model response `Wang Wei is an actor' based on the dialogues retrieved. Despite higher accuracy due to analogous character experiences, case 2 still provides an incorrect answer. The key difference between cases 2 and 3 is the memory classification mechanism. While case 2 retrieves relevant dialogues, it fails to retrieve essential semantic memory as in case 3. With memory classification, our models retrieve accurate social relationship memory, yielding correct responses. In this evaluation, with 'cameraman' as the memory anchor, only case 3 correctly incorporates the pertinent memory.

\section{Conclusion}
Our study introduces the PerLTQA dataset, which features a memory database and memory-based QA pairs. This dataset encompasses personal long-term memory, including profiles, social relationships, events, and dialogues, divided into semantic and episodic memory categories. We also outline three subtasks: memory classification, retrieval, and synthesis, and report on baseline experiments with five LLMs and three retrievers. Our findings reveal that Bert-based memory classification surpasses other LLMs in categorizing memory types. We also note considerable differences among LLMs in generating accurate memory-based answers. This research significantly deepens the understanding and evaluation of LLMs in the context of personal long-term memory.

\section*{Limitations}
In this work, we utilize \texttt{gpt-3.5-turbo} to generate a memory-based dataset and evaluate its ability to generate responses based on memory in three distinct subtasks. However, we acknowledge the following limitations: 1. The process of generating memory data in the PerLTQA memory database could be varied. We have only implemented a step-by-step generation method based on memory types. Furthermore, the prompts used during the generation process still have room for optimization. 2. All the content is fictional. Despite our thorough manual screening, there may still be instances where the common knowledge presented does not align with reality. 3. Our evaluations are limited to four open-source LLMs that are less than 10B in size and ChatGPT. We do not evaluate other LLMs of varying scales and types. 4. For the evaluation of the correctness and coherence of response generation, we adopted the evaluation methods of LLMs. However, this metric may still have uncertainties in accurately measuring the quality of responses.

\section*{Ethics Statement}
The work presented in this paper introduces the PerLTQA dataset, which is generated from ChatGPT (\texttt{gpt-3.5-turbo}). This dataset does not violate any licenses or policies, nor does it infringe on privacy. The dataset can be utilized for academic exploration in memory-based QA, dialogue, and other related fields. To ensure the quality of the data, we have employed three researchers in the field of natural language who are proficient in both Chinese and English and possess excellent communication skills. Each researcher is paid \$20 per hour (above the average local payment of similar jobs). The design, annotation, and review of the entire dataset took four months, costing approximately an average of about 200 hours per annotator. The annotators have no affiliation with any of the companies that are used as targets in the dataset, eliminating any potential bias due to conflict of interest.

\bibliography{anthology,custom}
\bibliographystyle{acl_natbib}

\clearpage
\appendix
\section{Appendix}
\label{sec: appendix}

\subsection{Memory Database Generation Prompts}
\label{memory_generation_prompt}
The design of the PerLT memory dataset prompts are illustrated in Figure \ref{fig5:variamem_prompt}. The "Profile Generation" prompt creates character profiles using specified seed data and a prompt template. Following this, the "SR (Social Relationship) Generator" prompt produces social relationships based on ten provided seed relationships. Additionally, the "EVT (Event) Generator" prompt is employed to create events that align with the established social relationships between characters. Lastly, the "DLG (Dialogue) Generator" prompt facilitates the generation of event-based dialogues between a character and an AI assistant. Collectively, these prompts enable our model to generate raw memory data effectively.

\begin{figure}[htbp]
\centering
\includegraphics[width=0.48\textwidth]{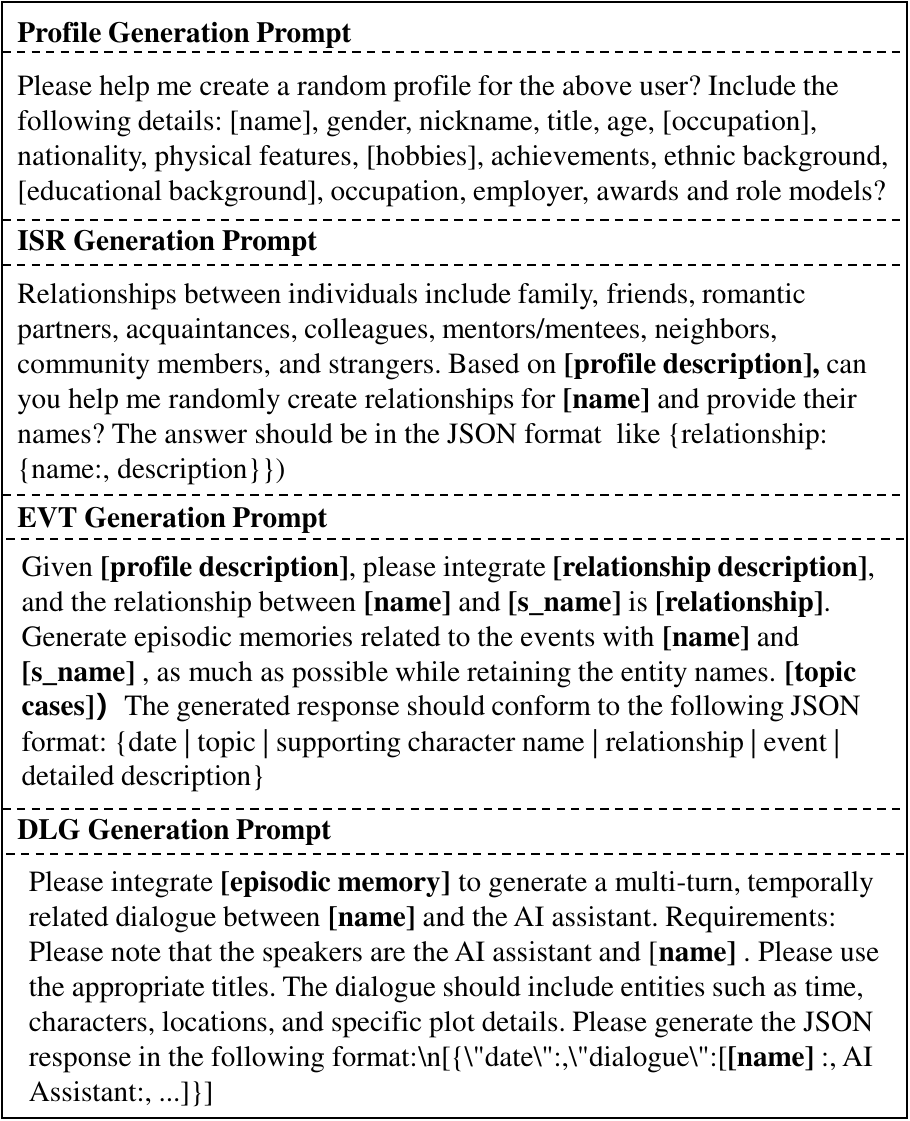}
\captionsetup{justification=justified}
\caption{Prompts for PRO, SR, EVT, and DLG memory generator.}
\label{fig5:chrono_prompt}
\vspace{-0.5cm}
\end{figure}

\subsection{Memory QA items Generation Prompts}
\label{memory_qa_prompt}
The design of the PerLT QA generation prompts are illustrated in Figure \ref{fig5:chrono_prompt}. The "Question and Answer Generation" prompt is designed to create questions and answers based on a provided reference memory and character name. Additionally, the "Memory Anchor Candidates Searching" prompt is utilized to identify key fragments that are crucial for crafting questions. These fragments are specifically chosen because they are present both in the generated answer and in the reference answer, ensuring relevance and coherence.

\begin{figure}[htbp]
\centering
\includegraphics[width=0.48\textwidth]{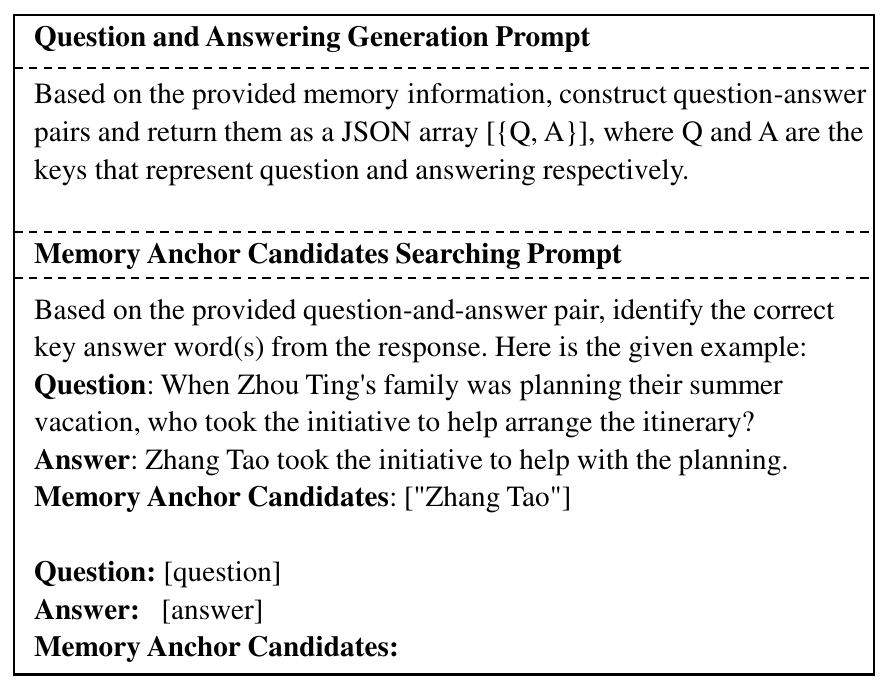}
\captionsetup{justification=justified}
\caption{Prompts for question answering generation, and memory anchor candidate searching.}
\label{fig5:variamem_prompt}
\vspace{-0.5cm}
\end{figure}

\subsection{Dataset Generation Error Types}
\label{memory_error_types}
In the dataset generation process for PerLT Memory and PerLT QA, several categories of errors are identified and corrected as shown in Table \ref{error_type_tab}. Anomalies, such as missing information in profiles, are rectified by removing or emptying the faulty fields. Incorrect character relationships that do not provide sufficient event data are excluded from the dataset. Instances of brief event narratives without detailed information are eliminated. Referent errors, which include incorrect or ambiguous references, are replaced with accurate information to ensure clarity. Redundant answers are streamlined to avoid unnecessary repetition, ensuring concise and relevant data. Finally, blurred memory anchor boundaries are corrected to precisely reflect the intended memory cues. These steps are taken to enhance the accuracy and reliability of the dataset.
\begin{table*}[t]
\centering
\renewcommand{\arraystretch}{1.05}
\resizebox{16cm}{!}{
\begin{tabular}{l|l|l|c|l}
\toprule
\textbf{Error Type} & \multicolumn{1}{c|}{\textbf{Source}} & \multicolumn{1}{c|}{\textbf{Error Example}} & \multicolumn{1}{l|}{\textbf{Operation}} & \multicolumn{1}{c}{\textbf{Revision}} \\ \hline
\begin{tabular}[c]{@{}l@{}}Anomalies \\ in profiles\end{tabular} & PerLT Memory & \{hobbies: “Not Provided”\} & Remove & \{hobbies: “”\} \\ \hline
\begin{tabular}[c]{@{}l@{}}Invalid\\ character relationship\end{tabular} & PerLT Memory & \begin{tabular}[c]{@{}l@{}}Zheng Yong has a wife and \\ girlfriend at the same time.\end{tabular} & Remove & \begin{tabular}[c]{@{}l@{}}Remove the relationship wife or girlfriend \\ which not provide enough events data.\end{tabular} \\ \hline
\begin{tabular}[c]{@{}l@{}}Brief\\ event narratives\end{tabular} & PerLT Memory & \begin{tabular}[c]{@{}l@{}}Xiaoming’s father used to \\ participate in the activities.\end{tabular} & Remove & - \\ \hline
Referent error & PerLT QA & \begin{tabular}[c]{@{}l@{}}When will Wang Xiaoming and the AI \\ assistant plan to visit the exhibition?\end{tabular} & Replace & \begin{tabular}[c]{@{}l@{}}When will Wang Xiaoming and Wang Xiaohong \\ plan to visit the exhibition?\end{tabular} \\ \hline
\begin{tabular}[c]{@{}l@{}}Redundant \\ answer\end{tabular} & PerLT QA & \begin{tabular}[c]{@{}l@{}}Who is the mentor of Wangxiaoming?\\ Wangxiaoming’s mentor is Zhangwen.\end{tabular} & Reduce & Zhangwen. \\ \hline
\begin{tabular}[c]{@{}l@{}}Blurred\\ Memory anchor boundaries\end{tabular} & PerLT QA & \begin{tabular}[c]{@{}l@{}}Answer: They met at Bali\\ Memory Anchor:{[}“At Bali”{]}\end{tabular} & Correct & \begin{tabular}[c]{@{}l@{}}Answer: They met at Bali\\ Memory Anchor:{[}“Bali”{]}\end{tabular} \\ \bottomrule
\end{tabular}
}
\captionsetup{justification=justified}
\caption{The error types observed in PerLT Memory and QA items generation and revision by human.}
\label{error_type_tab}
\end{table*}

\subsection{Optimizing Memory Retrieval with Memory Classification Re-Ranking}
\label{opt_rerank}
We devise a method in which the output probabilities of the classification model are utilized to furnish the retrieval model with classification insights, allowing for the re-ranking of candidate memories. This strategy minimizes the risks associated with memory retrieval based on specific memory bank classification results. Such risks primarily stem from potential classification inaccuracies that could lead to memory retrieval from an incorrect memory type, thereby unduly influencing the reliance on classification model precision within the framework. The introduction of a re-ranking strategy ensures the retrieval of a predefined number of memories across all memory types, regardless of the initial confidence levels of classification results. This is achieved through a weighted score re-ranking mechanism that effectively reduces the influence of classification inaccuracies on the ultimate ranking. For those instances with high classification confidence, revising their scores and reordering them accentuates their relevance, thereby optimizing the retrieval process.

\begin{figure}[htbp]
\centering
\includegraphics[width=0.48\textwidth]{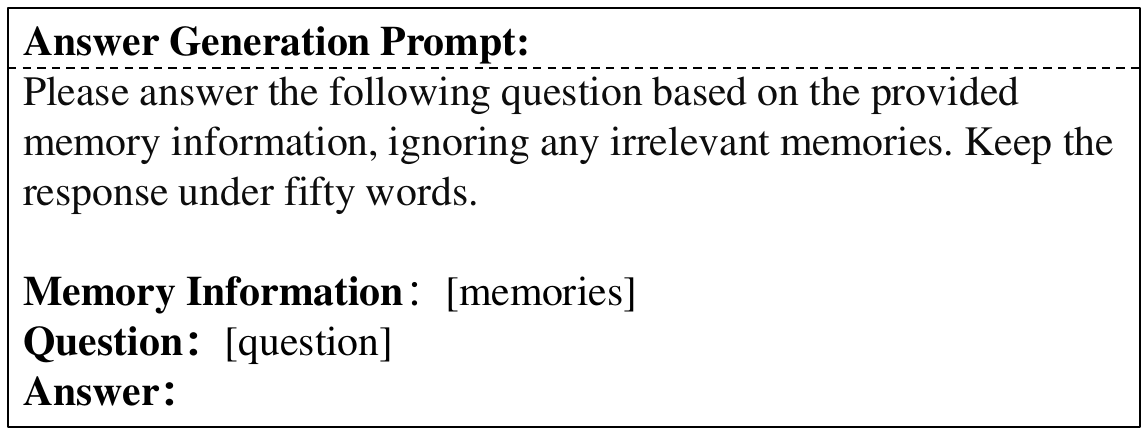}
\captionsetup{justification=justified}
\caption{Prompts for answering generation.}
\label{t_prompt}
\vspace{-0.5cm}
\end{figure}

\subsection{Experiment Settings}
\label{exp_settings}
\textbf{Memory Classification settings.} 
We conduct binary-class classification experiments on semantic memory, and episodic memory using BERT, Baichuan, ChatGLM2, ChatGLM3, and ChatGPT. For BERT, we employ fine-tuning with the evaluation questions to predict the memory type. For LLMs, we use instructions to guide LLMs in predicting the memory type. We also conduct instruction augmentation BERT experiments. Specifically, we train BERT-base classification models with 7,516 QA pairs. We finally evaluate the performance of memory type classification on a test set of 1,719 evaluation questions.

\textbf{Memory Retrieval settings.} We create unique memory banks for each character. In the case of DPR, we train the DPR model using 7516 evaluation questions. Contriever uses the text2vec model \citep{Text2vec} from Hugging Face to calculate the similarity between memory sentences and questions.

\textbf{Memory Synthesis settings.} In the W-MC+R setting, responses are generated using retrieved memories that are post-ranked based on memory classification outcomes. Conversely, in the W/o-MC+W+R scenario, responses are produced solely through memory retrieval, without the aid of memory classification for re-ranking. Meanwhile, in the W/o-MC+R framework, responses are generated directly without utilizing any external memory, relying solely on the inherent knowledge in LLMs. These configurations not only validate the effectiveness of each component but also underscore the importance of external memory. Due to limited resources, we only evaluated LLMs with fewer than 10 billion parameters. These models are prompted by retrieved memories. To ensure smooth operation on an Nvidia-3090 GPU with 24GB of memory, we have implemented a semi-precision inference setting.

\end{document}